\documentclass[sigplan,nonacm]{acmart}
\usepackage{enumitem}
\usepackage{graphicx}
\usepackage{subcaption}
\usepackage{geometry}
\usepackage{tabularx}
\usepackage{caption}
\usepackage{placeins} 

\AtBeginDocument{%
  \providecommand\BibTeX{{%
    \normalfont B\kern-0.5em{\scshape i\kern-0.25em b}\kern-0.8em\TeX}}}

\begin{document}

\title{Neural Appearance Modeling From Single Images}
\author{Jay Idema}
\email{jtidema@wm.edu}
\affiliation{%
  \institution{William \& Mary}
    \streetaddress{}
  \city{Williamsburg}
  \state{Virginia}
  \country{USA}
  \postcode{}
}
\author{Pieter Peers}
\email{ppeers@cs.wm.edu}
\affiliation{%
  \institution{William \& Mary}
    \streetaddress{}
  \city{Williamsburg}
  \state{Virginia}
  \country{USA}
  \postcode{}
}

\begin{teaserfigure}
  \resizebox{0.95\linewidth}{!}{
  \includegraphics{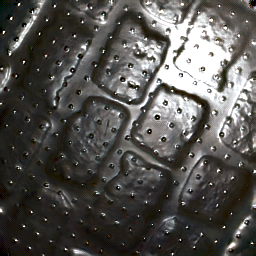}
  \includegraphics{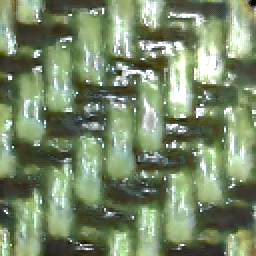}
  \includegraphics{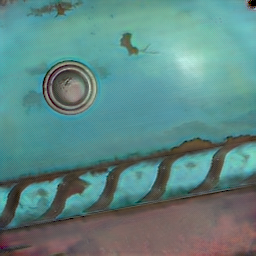}
  \includegraphics{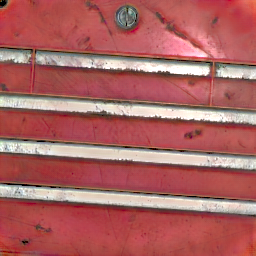}
  }
  \caption{Neural re-renders of input photographs. Appearance properties estimated from a single HDR photograph with co-located light and view, then re-visualized using a trained neural renderer.}
  \label{fig:teaser}
  \end{teaserfigure}

\begin{abstract}
We propose a material appearance modeling neural network for visualizing plausible, spatially-varying materials under diverse view and lighting conditions, utilizing only a single photograph of a material under co-located light and view as input for appearance estimation. Our neural architecture is composed of two network stages: a network that infers learned per-pixel neural parameters of a material from a single input photograph, and a network that renders the material utilizing these neural parameters, similar to a BRDF. We train our model on a set of 312,165 synthetic spatially-varying exemplars. Since our method infers learned neural parameters rather than analytical BRDF parameters, our method is capable of encoding anisotropic and global illumination (inter-pixel interaction) information into individual pixel parameters. We demonstrate our model's performance compared to prior work and demonstrate the feasibility of the render network as a BRDF by implementing it into the Mitsuba3 rendering engine. Finally, we briefly discuss the capability of neural parameters to encode global illumination information.

\end{abstract}


 \renewcommand\shortauthors{Idema and Peers}
 
\maketitle

\section{Introduction}

Plausibly modeling the appearance of spatially varying materials from a single photograph is a challenging problem. Perfect appearance reconstruction from a single photograph is notably under-constrained; an uninteresting solution that always perfectly recreates the initial photograph's appearance would be to assume it is a photograph of a completely non-glossy colored print-out of a prior photograph of the actual material, thus disregarding more complex observed reflectance properties. It is thus beneficial to seek to estimate plausible, 'interesting' underlying materials of single photographs, rather than finding a solution which best matches the input photograph.

In an optimistic best-case scenario for single image appearance estimation, the photograph is of a material which varies minimally in reflectance properties across its surface, and provides comprehensive data that accurately captures both diffuse and specular properties of the material's reflectance properties. In other words, the image contains information about how light interacts with the material both in the case of mirror-like reflections (specular effects) and scattered reflections (diffuse effects). Assuming the spatially-varying reflective properties of the material vary predictably across the surface and can thus be correlated, an estimation method could fit material parameters to multiple data points (multiple pixels in a single photograph), where the lighting and viewing conditions differ. In reality, this is far too optimistic and several primary challenges must be addressed to fully utilize a single photograph for appearance estimation. One primary issue is accurately segmenting the material surface into regions representing fundamentally different reflectance properties, such as distinguishing diffuse painted areas from glossy metallic areas. This requires techniques capable of recognizing diverse material types within the same photograph, a challenge further exacerbated by pixels that may capture and blend multiple sub-materials of a surface. Additionally, understanding how these 'macro' sub-materials of a surface are locally perturbed, such as by dents in metal or variations in paint thickness, necessitates methods that can infer (or hallucinate) fine-scale surface details from limited visual information.

Most existing appearance modeling solutions attempt to estimate the parameters of an analytical Bidirectional Reflectance Distribution Function (BRDF). A BRDF is a function which takes an incident light direction and outgoing view direction as input and returns the ratio of incident irradiance to outgoing radiance for a single point on a surface. Spatially Varying Bidirectional Reflectance Distribution Functions (SVBRDFs) model the ratio of reflected radiance to incident irradiance across different points on a surface, capturing variations in reflectance properties across the surface. Common isotropic BRDFs like GGX are modeled as SVBRDFs by parameterizing a material surface based on each surface position's diffuse reflectance (scattering light uniformly) and specular reflectance (scattering light in a specific mirror directions based on microfacet perturbations). Within the parameter space of SVBRDFs, different SVBRDF parameter maps for a material surface may result in the same visual appearance under the same lighting conditions. Many single image appearance modeling methods, including ours, attempt to produce plausible, 'interesting' material appearances from known lighting, rather than truly reproducing a surface's underlying material properties or perfectly recreating the input photograph \cite{bieron2023single}.

In recent years, machine learning has driven significant advancements in successful techniques for single image appearance modeling. Most of these techniques center on the 'estimation' of per-pixel analytical (derived) SVBRDF parameter maps \cite{deschaintre2018single, guo2021highlight, zhou2021adversarial}. In addition to suffering from the aforementioned under-constrained nature of single image appearance modeling, analytical SVBRDFs also do not perfectly model all the complex reflectance properties of real surfaces \cite{guarnera2016brdf} nor do they directly capture complex inter-pixel light transport effects like shadows. In contrast to direct parameter inference methods, neural re-rendering methods seek to re-visualize photographed materials without explicit analytical parameter estimation, instead utilizing learned neural networks to estimate the visual appearance of a photographed material without explicit analytical BRDF parameters. These methods are thus wholly dependent on the training data to determine the reflectance effects which they model \cite{bieron2023single}. Existing neural re-rendering methods relight whole images, making them unsuitable and inefficient for path-tracing rendering engines, which require per-pixel reflectance evaluation for efficient use.

We propose a neural re-rendering method which generates 'neural material parameters' at a per-pixel granularity. Unlike rigid analytical SVBRDF parameters, these parameters are learned during training alongside a neural renderer which is trained to utilize these learned parameters for (re-)rendering. Inspired by work in Neural BRDFs (NBRDFs) and neural Apparent-SVBRDF compression \cite{zhou2023photomat, rodriguez2023neubtf, bieron2023single}, we jointly train two neural networks. Together, the full architecture is trained to perform image relighting, transforming an input photograph of a planar material exemplar into a re-visualization of the photographed material under different light and view conditions. The first and larger network transforms an input photograph into a spatially varying map of view-and-light-independent neural material parameters (or SV-NBRDF parameters) with equal resolution to the input photograph. The second, smaller network takes a per-pixel neural parameter vector, the incoming light position, and outgoing view position as inputs to generate an RGB output color. This smaller network is shared across all pixels and acts as a function of light and view directions parameterized by a pixel's estimated neural material parameters.

Importantly, our model's second network returns the ratio of the reflected radiance to the incident irradiance, multiplied by the projected area of the pixel with respect to the incident light direction. It is thus easily refactored into a BRDF by dividing by the projected area of the pixel with respect to the incident light direction. This property of the network and the two-stage modularity of our architecture allows our neural rendering network to be efficiently implemented into rendering engines at the pixel granularity. We envision that our neural BRDF parameters can further be made editable \cite{hu2020deepbrdf}, just as analytical BRDF parameters are.\\
In conclusion, our primary contributions are:
\begin{enumerate}[topsep=0pt]
    \item A novel two-stage neural network strategy for plausible appearance modeling from a single photograph;
    \item a demonstration of our method's potential to encode and model global illumination effects like shadows via its parameters, which is not possible by naively using analytical BRDFs; and
    \item a demonstration of the feasibility of implementing our neural renderer into the path-tracing rendering engine, Mitsuba3 \cite{Mitsuba3}.
\end{enumerate}
\ \\
Code for this project, as well as a simple Jupyter Notebook demo, can be found at \href{https://github.com/APeculiarCamber/neural_appearance_estimation}{https://github.com/APeculiarCamber/\\neural\_appearance\_estimation}.

\section{Related Work}
\textbf{Single Image SVBRDF Estimation.} Due to its under-const-rained nature, inferring SVBRDF parameters from a single photograph is a very challenging problem. Nascent machine learning work demonstrated the feasibility of neural network solutions for inferring SVBRDF parameters from a single photograph; Li et al. \cite{li2017modeling} utilize a convolutional neural network to infer the GGX SVBRDF parameters of unlabelled single images under natural lighting, for a small category of materials. They illustrated that the problem could be classified as the inverse of the rendering algorithm and highlighted the capability of neural networks to address the inherent ambiguities in estimating single image appearance.

Subsequent research has focused on guiding neural networks to more plausibly resolve ambiguities, particular ambiguities caused by specular highlights and lossy low dynamic range (LDR) image clamping. Deschaintre et al. \cite{deschaintre2018single} compile a dataset of over 200,000 synthetic SVBRDFs to train a network on sufficient materials such that appearance ambiguities are resolved plausibly by the model; they further proposed using a rendering loss as a alternative similarity metric compared to a direct SVBRDF parameter comparison loss. To manage over-saturated highlights in LDR images, Guo et al. \cite{guo2021highlight} utilize highlight-aware (HA) convolutions, which operate similar to gated convolutions \cite{yu2019free}. Alternatively, other approaches seek optimization-based solutions. Li et al. \cite{li2018materials} utilize a ‘refinement’ step to combat over-saturation. Zhou and Kalantari \cite{zhou2022look} perform test-time optimization on estimated SVBRDF parameters to minimize test error. MatFusion \cite{sartor2023matfusion} utilizes an iterative diffusion model to synthesize multiple SVBRDF parameter estimations, allowing for user or heuristic selection between multiple parameter estimations. MatFusion also augments Deschaintre et al.’s synthetic SVBRDF dataset \cite{deschaintre2018single} with mask-based material parameter map blending.

Other similar work has attempted to improve finer-detail quality, which poses a fundamental issue for neural networks and their tendency to prioritize low-frequency signals. Neural discriminator strategies (which jointly train discriminator networks to distinguish ground truth renders from neural renders) appear particularly effective at encouraging finer detail. ‘SurfaceNet’ \cite{vecchio2021surfacenet} is a GAN-based network trained using a neural discriminator loss. Zhou and Kalantari \cite{zhou2021adversarial} use per-channel neural discriminator networks on both real and synthetic materials. Focusing on fidelity via higher resolution, other methods \cite{martin2022materia, guo2023ultra} have been proposed to estimate SVBRDFs for 'ultra' high resolution photographs, using stitching techniques to process smaller tiles of high resolution photographs.

These advancements have been instrumental in improving the quality of appearance estimation. Notably, our training process harnesses the Matfusion dataset \cite{sartor2023matfusion} augmented from Deschaintre et al.'s dataset \cite{deschaintre2018single}; and our architecture utilizes highlight-aware convolutions \cite{guo2021highlight}. However, these methods focus on directly inferring analytical SVBRDF parameters. Despite their ubiquity, these rigid analytical parameters exhibit limitations in their capacity to represent intricate anisotropic reflectance characteristics and to incorporate inter-pixel reflectance interactions, such as material-wide light transport (global illumination). Our method, when provided with training data featuring these complexities, can theoretically remain robust.

\textbf{Neural Appearance Modeling.}
Single image neural appearance modeling methods forego inferring analytical SVBRDF parameter maps and instead investigate neural parameter spaces. PhotoMat \cite{zhou2023photomat} is a method for fine-tuning a generative model on single images to construct spatially varying implicit neural materials; these implicit neural materials are rendered per-pixel by a jointly trained conditional neural renderer. Notably, PhotoMat is not a inverse rendering method, but instead a generative model which uses a single image for tuning material generation. Within the domain of inverse rendering, Bieron et al. \cite{bieron2023single} propose injecting lighting information into the decoder of a U-Net to relight single images directly without implicit or explicit per-pixel material parameters. As the decoder is convolutional, it has no per-pixel parameter map, neither neural nor analytical.

Our work utilizes a similar encoder-decoder model to Bieron et al.'s, but instead of a relit image, we train our U-Net to output neural parameter maps similar to PhotoMat’s. This greatly increases the efficiency and usability of our model over both PhotoMat (which only generates similar materials via a fine-tuning process) and Bieron et al.'s Neural Relighting method (which is only able to relight an image in full, requiring processing of its entire decoder for each and every desired light and view conditions).

\textbf{NBRDFs.}
Neural networks further offer a novel range of methods for encoding material properties beyond analytical SVBRDFs. Most research on neural representations of reflectance properties centers on compression and interpolation of Apparent BRDFs (ABRDFs). ABRDFs are measurement-based representations, encapsulating the reflective characteristics of a materials through extensive capture of photographs of a material surface for exhaustively many light and view conditions. DeepBRDF \cite{hu2020deepbrdf} is a direct neural compression method using an encoder-decoder architecture to encoding and decoding full ABRDFs. Rainer et al. \cite{rainer2019neural} use an asymmetric encoder-decoder in which a sorted ABRDF is compressed into an implicit neural feature vector. These neural features are rendered by the decoder when coupled with injected light and view information, returning an RGB color. Most similar to our work, NeuBTF trains a two-stage encoder-decoder model to infer a map of implicit neural material parameters alongside a renderer MLP \cite{rodriguez2023neubtf}. However, this model is trained on individual materials and input photographs only serve to augment appearance within the limited bounds of its trained single material. 

Although NeuBTF shares clear architectural similarities to our model, none of the related neural material representation methods perform single image appearance estimation. Instead, they all necessitate a complete ABRDF for training and/or compression. Our method is trained on hundreds of thousands of different spatially-varying materials; and our method does not require full ABRDFs for training nor for appearance estimation.

\section{Method}

\subsection{General Overview}
Our method takes as input a single photograph of a near-planar material surface, captured from known co-located view and light positions, and outputs per-pixel neural parameters which encode the appearance properties of the material. These parameters can be used to re-visualize the material from novel light and view positions, including non-co-located light and view configurations, using a neural renderer. Both input photographs and output visualizations are high-dynamic range (HDR) images, which capture a wider range of brightness levels than many camera sensors are capable of processing as distinct brightness levels; in contrast to low dynamic range (LDR) images which clamp high reflectance values to a range that most camera sensors can record as distinct. Common image file formats like Portable Network Graphics (PNGs) store photographs in a clamped low dynamic range.

All visualizations of material surfaces are perspective-rectified to ensure the material visualizations appear as if viewed from the top down, regardless of the actual viewing direction, which may be from any direction. In other words, perspective effects from the view direction are not applied to visualizations. For rendering purposes, we achieve perspective rectification by treating each pixel of a spatially varying surface as an independent BRDF at varied positions between the surface bounds of (-1,-1) for the 'top-left' corner and (1,1) for the 'bottom-right' corner. For a 256x256 image example: given a matrix of the positions of the image's pixels \(\mathbf{P} \in \mathbb{R}^{256 \times 256 \times 3}\) and a source point light \(\mathbf{L} \in \mathbb{R}^3\), the matrix of normalized light directions to each pixel from the light source \(\mathbf{\omega_i} \in \mathbb{R}^{256 \times 256 \times 3}\) can be computed as follows:
\[
\mathbf{\omega_i}^{xy} = \frac{\mathbf{T} - \mathbf{P}_{xy}}{\|\mathbf{T} - \mathbf{P}_{xy}\|}
\label{eq:position}
\] where \(\mathbf{P}_{xy}\) represents the position vector at the \((x,y)\) entry of \(\mathbf{P}\), and \(\|\cdot\|\) denotes the Euclidean norm. The view directions ($\omega_o$) can be similar computed. This results in perspective-rectified light and view directions per-pixel without foreshortening (visual distortion that occurs when a surface is viewed from an angle rather than straight on, causing it to appear compressed along the line of sight). Like an analytical BRDF, our neural renderer is thus a function only of material parameters, incident light direction, and outgoing view direction; not light and view positions.

\subsection{Radiometric Compression}
All synthetic (possessing artificial SVBRDF parameter maps) materials are rendered using the isotropic GGX BRDF model \cite{walter2007microfacet, trowbridge1975average} in linear RGB color-space.

In order to avoid large values dominating the model and its losses during training (and to encourage training to prioritize the more visually importance lower brightness range [0, 1]), we further apply a simple, reversible log-scaling function to all reflectance values, R, during processing:
\[
\mathbf{R}_{log} = \log (R + 1)
\]
In order to accurately display a visualization, it must be converted out of log-scale space and then out of linear RGB color-space, by applying a gamma correction of 2.2:
\[
\mathbf{R}_{vis} = (e^{R_{log}} - 1)^{\frac{1}{2.2}}
\] $R_{vis}$ can then be rendered to a screen at a perceptually appropriate brightness.

\subsection{Input Photographs}
Our method assumes input photographs are captured by a camera with an FOV of $28^\circ$. We train only on input photographs lit and viewed from positions directly above the center of the surface, looking directly towards it. This greatly simplifies input considerations, allowing the model to focus on plausible appearance modeling. For co-located light and view positions, this also creates a large, central specular highlight, vital for determining the specular reflectance properties of the input material. We intentionally select a narrow FOV of $28^\circ$ as photographs with wider FOVs can be easily cropped to narrower FOVs, while the inverse is more difficult.

\begin{figure*}[!t] 
\centering
    \includegraphics[width=0.92\textwidth]{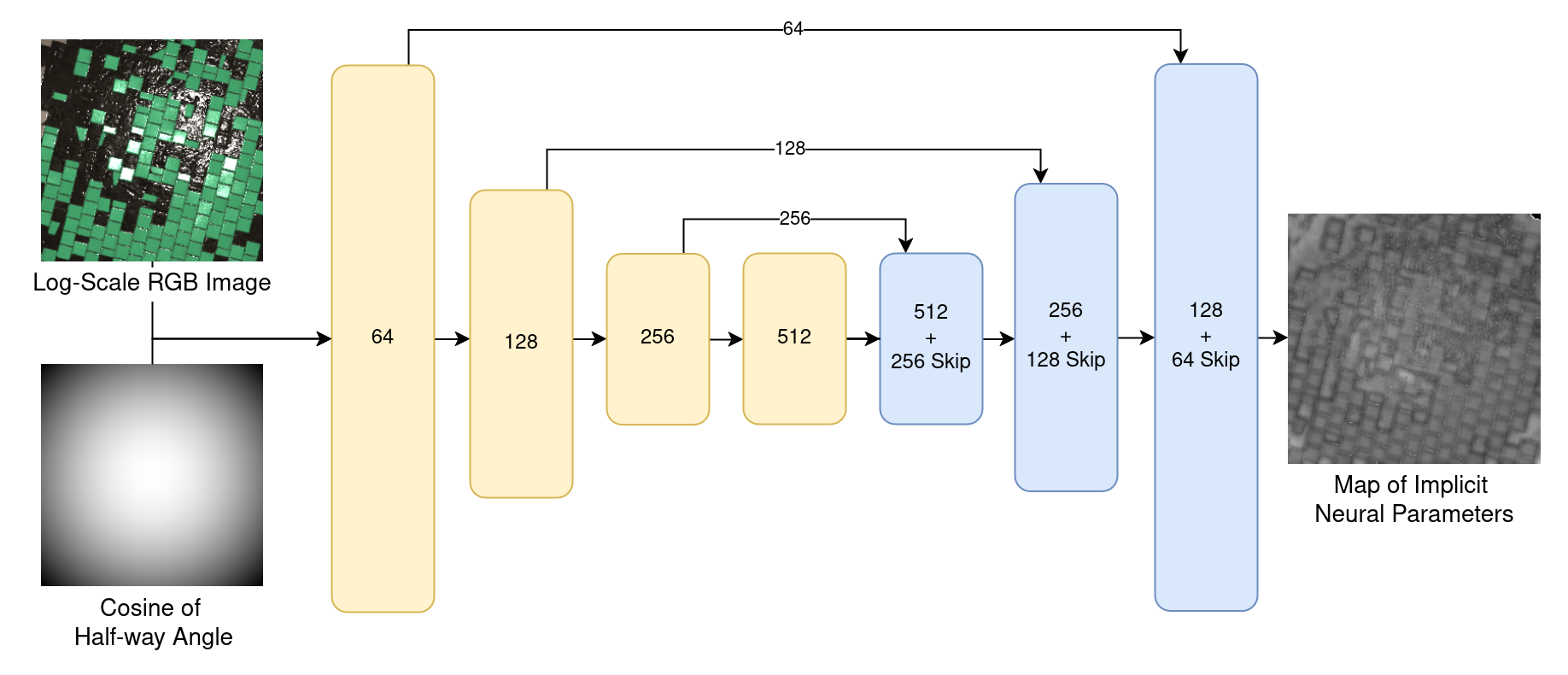} 
    \caption{The first stage of our architecture, $M_{\text{est}}$, which transforms an input RGB image and map of half-direction cosines into a map of implicit neural material parameters, to be rendered by $M_{\text{render}}$. The yellow layers represent Highlight-Aware residual Convolution blocks of 2 HA convolutions each. The blue layers represent standard residual convolution blocks.}
    \label{fig:U-Net}
\end{figure*}

\subsection{U-Net}
The initial stage of our network, $M_{\text{est}}$, takes a photograph of a material under co-located view and light conditions and transforms it into an equal-resolution 2D map consisting of length-64 parameter vectors (referred to as 'implicit' neural material parameters). In order to encode the light and view information into the input for better generalization to perturbations in camera position, we concatenate each log-relative RGB vector of each pixel with the cosine of the angle between the macro-surface normal, n, and the half-way direction, $\omega_h$, for that pixel. The half-way direction, $\omega_h$, is the normalized sum of the incident light direction $\omega_i$ and the view direction $\omega_o$, i.e. ($\omega_h$ = ($\omega_i$ + $\omega_o$) / $||\omega_i$ + $\omega_o||$). This creates an input tensor of four channels (R, G, B, ($\omega_h \cdot n$)), where $\cdot$ is the dot product of the vectors. Log-scaling is not applied to the cosine of the half-way angle. To determine n, we ignore potentially unknown spatially-varying perturbations in surface orientation and assume n is the macro-surface normal of the entire material, $n=(0, 0, 1)$. Thus, ($\omega_h \cdot n$) is equivalent to the z component of $\omega_h$.

An encoder-decoder U-Net is chosen for $M_{\text{est}}$ due to a U-Net's desirable capabilities for image segmentation and distribution learning \cite{ronneberger2015u}. The architecture of $M_{\text{est}}$ is directly adapted from an U-Net architecture used for single image material relighting by Bieron et al. \cite{bieron2023single}.  The encoder begins with a single 7x7 kernel convolution to expand the four input channels to 64 channels; a typical structure of residual block convolutions \cite{he2016deep} then follows. Downsampling is performed thrice by 2x2 kernel convolutions. 

The convolution layers of the encoder's residual blocks are replaced with highlight aware (HA) convolution layers \cite{guo2021highlight} to mitigate ‘burn-in’ effects, in which overexposed highlight regions of photographs result in poor appearance estimation. Specifically, these 'burn-in' effects manifest as methods under- or over-estimating diffuse reflectance in highly bright specular regions. Real-world, LDR photographs often have overexposed highlight regions where radiance near the specular highlight exceed the maximum for low dynamic range images and are thus clamped. These overexposed regions correspond to regions in HDR photographs where the radiance values are higher than the maximum radiance of LDR photographs and are thus clamped to the lower range when converting to LDR; this results in an unrecoverable loss of information. The loss of information for the true reflectance behaviors of materials photographed or saved in low dynamic range (LDR) causes brighness ambiguities, as all brightness values above the LDR upper bound are clamped to the same value. HA layers seek to mitigate the ambuiguities this causes by utilizing a method similar to gated convolutions used in image in-painting \cite{yu2019free}. Unlike gated convolutions however, HA layers learn to apply a dynamic mask based on learned identification of over-exposed regions, rather than processing a predefined in-paint mask.

Lastly, the decoder of $M_{\text{est}}$ uses residual blocks with standard convolutions, under the assumption that the encoder will sufficiently mask 'burn-in' inducing highlights. Upsampling in the decoder is performed by 4x4 transposed convolutions. Skip connections between the encoder and decoder are added after each of the three upsamples in the spatial dimensions. The final 'decoded' output is a 64-channel map of implicit neural parameters with the same spatial dimensions as the input photograph. See Figure \ref{fig:U-Net} for a diagram of this simple network.

\subsection{Renderer}
A BRDF is a function which takes an incident light direction and outgoing view direction and returns the ratio of incident irradiance to outgoing radiance. The second stage of our network, $M_{\text{render}}$, is functionally a parameterized BRDF model with radiance intensity foreshortening applied (decrease in reflected area from viewing at an angle). $M_{\text{render}}$ takes as input each 64-length implicit material parameter vector generated by $M_{\text{est}}$ and any pair of light and view directions, and generates as output a log-relative, linear-RGB color representing the appearance of the pixel's material (corresponding to its neural parameters) under the conditions of that input light and view direction.

While neural networks are highly effective at approximating low frequency functions, they often struggle with high frequency approximation, such as the sharp transition into a specular highlights of smooth, shiny materials. To mitigate this, we generate positional encodings for the x,y,z components of the incident light direction $\omega_i$, output view direction $\omega_o$, and the half-way direction, $\omega_h$, at exponentially increasing frequencies; this technique allows high frequency information to be encoded in additional dimensions and is directly inspired by NeRF's positional encodings \cite{mildenhall2021nerf}. 
Formally, we encode 3D directions to a higher-dimensional space using a set of sinusoidal varied-frequency functions. Given an input direction (x, y, z) and number of frequencies n, the position encoding \(\Gamma(\mathbf{x, y, z})\) is defined as:
\[
\gamma_k(\mathbf{x}) = \left( \sin(2^k \pi x), \cos(2^k \pi x) \right)
\]
\[
\Gamma(x,y,z) = (x,y,z) \cup \bigcup_{k=0}^n \left( \gamma_k(\mathbf{x}) \cup \gamma_k(\mathbf{y}) \cup \gamma_k(\mathbf{z}) \right)
\]

where the direction (x, y, z) are the directions computed by Equation \ref{eq:position}.

These encoded directions are processed through an auxiliary MLP network, $ND_{enc}$, in order to re-compress the information. This compressed sinusoidal encoding is concatenated with the implicit material parameters of a material’s neural representation to compose the true input to $M_{\text{render}}$. By trial and error, we choose a frequency count n of 16, resulting in 288 initial values (2 * 16 sinusoidal encodings, of three directions with three components each); and we choose a compressed encoding size of 32.

\begin{figure}[htbp]
    \centering
    \includegraphics[width=0.51\textwidth]{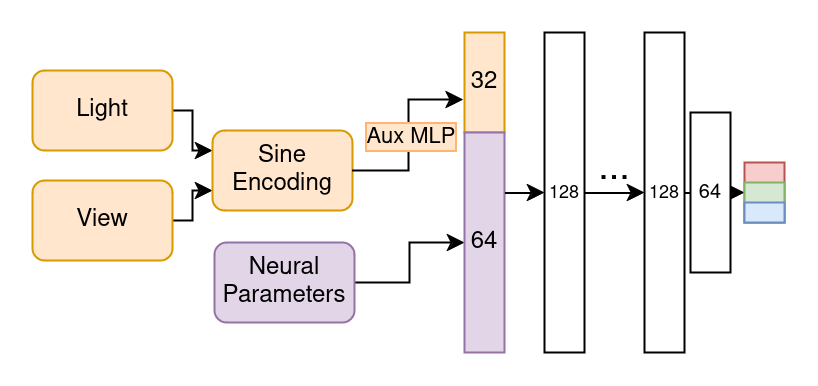} 
    \caption{MLP Per-pixel Neural Renderer (NBRDF). The neural parameters (purple) are concatenated with the MLP-compressed encoding of light and view directions (orange); this input is fed into a 6-layer MLP to return a log-relative linear-RGB color (far right).}
    \label{fig:renderer}
\end{figure}

For the architecture of $M_{\text{render}}$, we experimented with different neural renderer architectures, such as Siren MLPs \cite{sitzmann2020implicit}, however we empirically found that a simple 6-layer fully connected MLP with LeakyRelu activations worked best. This can trivially be implemented as a convolutional network where all convolutions have a kernel size of 1 to enable implicit processing of 2D images. Figure \ref{fig:renderer} shows a diagram of our $M_{\text{render}}$ and the process of light and view encoding.

\subsection{Training}

\textbf{Architecture.}\ $M_{\text{render}}$ is considerably smaller than $M_{\text{est}}$. $M_{\text{est}}$ possesses 84,008,960 parameters compared to $M_{\text{render}}$'s 387,974 parameters. As a result, the encoder-decoder constitutes both the primary execution and memory bottleneck for our architecture. The number of input photographs that can be accommodated in a single batch of training is greatly limited by the size of $M_{\text{est}}$, but the number of output re-visualizations is not so constrained by $M_{\text{render}}$ and its auxiliary $ND_{enc}$ network. Since $M_{\text{est}}$ produces light and view independent representations of the materials, we can take advantage of the smaller size of $M_{\text{render}}$ to visualize each estimated implicit neural material with multiple light and view conditions, allowing for rendering many more output images than our network could accept input images, greatly increasing the reliability of training.

We jointing train $M_{\text{render}}$, $M_{\text{est}}$, and $ND_{enc}$ using the MatFusion's synthetic BRDF dataset \cite{sartor2023matfusion}. This dataset is an extension of the INRIA synthetic BRDF dataset \cite{deschaintre2018single} with mask-based SVBRDF combinations; it is comprised of 312,165 GGX SVBRDF parameter maps of near-planar mixed and combined synthetic material surfaces, along with normal maps to approximate surface angle perturbations. Further, we employ a test set of 50 unique materials not part of the training set nor its material combination processes.

\textbf{Generating Training Batch.}\ All synthetic materials for training are rendered using the isotropic GGX BRDF model \cite{walter2007microfacet, trowbridge1975average}. For a single batch of training, we randomly select two material from the MatFusion dataset \cite{sartor2023matfusion}, and generate one RGB input render and a fixed number of RGB output rendered exemplars for each material. Input renders are always rendered with co-located light and view positions, positioned directly downwards towards the center of the material such that a camera with an FOV of $28^\circ$ fully captures the material. Assuming a 2x2 unit material, this means positioning the light and view at a distance $d_{\text{view}}$ from the center of the material where:
\[
d_{\text{view}} = \frac{1.0}{\tan(\frac{28}{2})} \approx 4.010781
\]

For output exemplars, the SVBRDF parameters are not used directly to train our model, only to render training exemplars. Thus, it is critical that training provide as many challenging rendered exemplars as possible. As specular highlights are both a desirable visual effect and are high-frequency in the spatial dimensions of the photograph, most training output exemplars should include a specular highlight (in cases where the material is not entirely diffuse). As such, output exemplar renders are generated identically to Bieron et al.'s training process \cite{bieron2023single}. 
A point, p, is uniformly selected on the 2x2 unit surface of the material; this point is then perturbed by a sample from a Gaussian normal distribution with mean 0 and standard deviation 2.
\[
p_x = (\xi_1 * 2) - 1 + \sigma(\xi_2, \text{mean}=0, \text{std}=2)
\]\[
p_y = (\xi_3 * 2) - 1 + \sigma(\xi_4, \text{mean}=0, \text{std}=2)
\] where each $\xi$ is a uniformly sampled value in [0, 1] and $\sigma$ is the Gaussian normal distribution. A view position, $v$ is then sampled uniformly from the hemisphere of radius $d_{\text{view}}$, centered on the surface. Finally, a light position, $i$, is selected that achieves a specular highlight centered at point p, i.e. such that:
\[
    \frac{i - p}{||i - p||} = reflect_p(\frac{v - p}{||v - p||})
\] where $reflect_p$ returns the mirror reflected direction of the input direction at point p; and $||x||$ is the magnitude of a vector x. Finally, the light position's distance from the point p, ${||
i - p||}$, is sampled by:
\[
||i - p|| = |\sigma(\xi_i, \text{mean}=0,\text{std}=2)| + 0.5
\]
 
\textbf{Losses.}\ We use a combination of 3 primary losses during training: a data loss $L_d$, a perceptual loss $L_p$, and a conditional discriminator loss $L_c$. Our data loss, $L_d$ is the L1 error between a output exemplar image and the model-rendered image for the same light and view position. While log-based losses have been shown to be effective for parameter fitting tasks with intense specular lobes, our model already operates fully in log-relative space, rendering a log-based loss less useful. 

$L_d$  directly trains the model towards the GGX-based appearance of the training set. Unfortunately, the per-pixel nature of the loss biases the model towards blurring specular highlights to mitigate high error from the sharp shift in intensity on a highlight's edge. We address this using two additional losses. A 'Learned Perceptual Image Patch Similarity' (LPIPS) metric \cite{zhang2018unreasonable} grounds the model towards visually plausible renders, while a neural discriminator loss determines if the neural-rendered image is of the same 'material' as the input image. The neural discriminator is a small convolutional network of six layers of kernel size 4; it is jointly trained alongside the main rendering model and empirically effective at detecting and penalizing blurring. By trial and error, we found that the best weighting for these losses were:
\[
L = 1.0 * L_d + 0.01 * L_p + 0.03 * L_d
\]
\textbf{Masking.}\ Unfortunately, two key training problems arise due to our architecture. First, $M_{\text{est}}$ is prone to instabilities, likely due to the masking behavior of highlight-aware convolutions. These instabilities result in seemingly random, small patches in the implicit material map which are comprised of high-norm, high-noise 'garbage' parameter vectors. Without intervention, $M_{\text{render}}$ learns to adapt to these nonsense parameters with average expected values, obscuring the loss that ought to be applied to such parameters. Second, $M_{\text{render}}$ is a shared, per-pixel renderer; its parameter training behavior is thus highly affected by the overall appearance of materials (which tends towards 0 for highly specular materials where only a specular highlight is bright on an otherwise dark surface). To mitigate these issues, we employ one of two pixel sampling methods to select only specific pixels to contribute to the data loss for training of $M_{\text{render}}$. We propose two methods to generate a multi-nomial (multivariate) probability distribution to sample pixels:
\begin{enumerate}
    \item The inverse norm of the neural parameters of each pixel;
    \item The squared norm of the RGB vector of each pixel of output exemplars.
\end{enumerate}
We sample these distributions for 60\% of the pixels, and use only those sampled pixels for the data loss of training $M_{\text{render}}$. The first method samples aggressively away from unstable neural parameters, while the second samples reasonably towards specular and high-intensity colors. Our models are trained on the first probability distribution, generated by channel-wise neural parameter norms.

\section{Results}
\textbf{Implementation.} We implemented and trained our neural network model in PyTorch. We used as starting learning rate of $10^{-4}$ for $M_{\text{est}}$, $M_{\text{render}}$, and $ND_{enc}$. We employ a learning rate decay of 1.5$\%$ every epoch of 20,000 batches and utilize the Adam optimizer \cite{kingma2014adam}. We used the full 312,165 SVBRDF training set (cropped to 192 × 192 resolution). Only 60$\%$ of pixels for each material are used to inform the data loss of $M_{\text{render}}$. Each batch is comprised of two uniformly sampled materials from the MatFusion dataset, with one input photograph and eight rendered output exemplars each. We train for 30 epochs at 192 x 192 resolution on a single Nvidia RTX A5000 for around $60$ hours. 

To bring our model up to processing the desired resolution of 256 x 256 images, we fine-tune our model with a learning rate of $10^{-4}$ and rate decay of $0.8$ every 20,000 batches. We finetune for 200,000 batches at 256 x 256 resolution for 24 hours; these 256 x 256 batches still use two input training materials but only render four output exemplars per material due to memory constraints.\\

\begin{figure*}[htbp]
\begin{subfigure}[t]{0.150000000\textwidth}
    \centering
    Input
    \vspace{0.800em}
\end{subfigure}
\begin{subfigure}[t]{0.150000000\textwidth}
    \centering
    Reference
    \vspace{0.800em}
\end{subfigure}
\begin{subfigure}[t]{0.150000000\textwidth}
    \centering
    Ours
    \vspace{0.800em}
\end{subfigure}
\begin{subfigure}[t]{0.150000000\textwidth}
    \centering
    Zhou 21
    \vspace{0.800em}
\end{subfigure}
\begin{subfigure}[t]{0.150000000\textwidth}
    \centering
    MatFusion
    \vspace{0.800em}
\end{subfigure}
\begin{subfigure}[t]{0.150000000\textwidth}
    \centering
    Bieron 23 - Relit
    \vspace{0.800em}
\end{subfigure}
\centering
\begin{subfigure}[b]{0.150000000\textwidth}
    \centering
    \includegraphics[width=\textwidth]{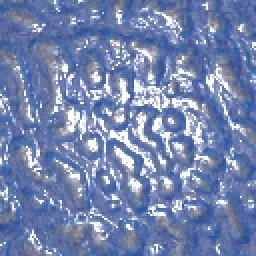}
\end{subfigure}
\begin{subfigure}[b]{0.150000000\textwidth}
    \centering
    \includegraphics[width=\textwidth]{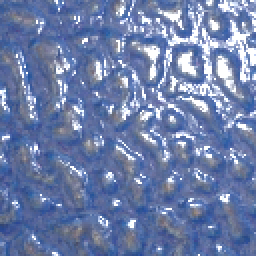}
\end{subfigure}
\begin{subfigure}[b]{0.150000000\textwidth}
    \centering
    \includegraphics[width=\textwidth]{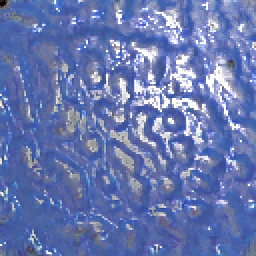}
\end{subfigure}
\begin{subfigure}[b]{0.150000000\textwidth}
    \centering
    \includegraphics[width=\textwidth]{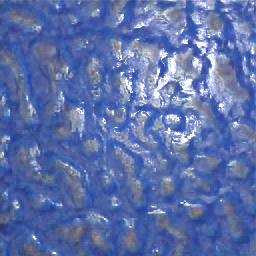}
\end{subfigure}
\begin{subfigure}[b]{0.150000000\textwidth}
    \centering
    \includegraphics[width=\textwidth]{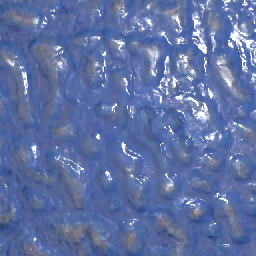}
\end{subfigure}
\begin{subfigure}[b]{0.150000000\textwidth}
    \centering
    \includegraphics[width=\textwidth]{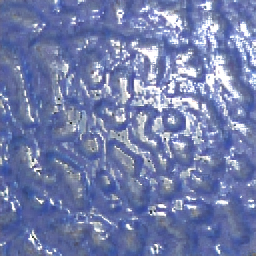}
\end{subfigure}
\centering
\begin{subfigure}[b]{0.150000000\textwidth}
    \centering
    \includegraphics[width=\textwidth]{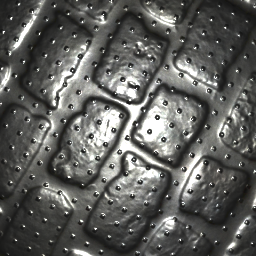}
\end{subfigure}
\begin{subfigure}[b]{0.150000000\textwidth}
    \centering
    \includegraphics[width=\textwidth]{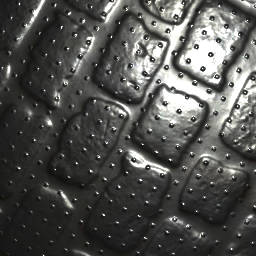}
\end{subfigure}
\begin{subfigure}[b]{0.150000000\textwidth}
    \centering
    \includegraphics[width=\textwidth]{results/ximage_1_row_2.png}
\end{subfigure}
\begin{subfigure}[b]{0.150000000\textwidth}
    \centering
    \includegraphics[width=\textwidth]{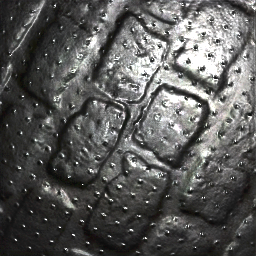}
\end{subfigure}
\begin{subfigure}[b]{0.150000000\textwidth}
    \centering
    \includegraphics[width=\textwidth]{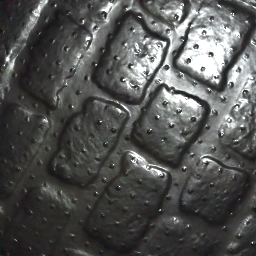}
\end{subfigure}
\begin{subfigure}[b]{0.150000000\textwidth}
    \centering
    \includegraphics[width=\textwidth]{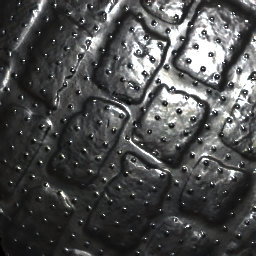}
\end{subfigure}
\centering
\begin{subfigure}[b]{0.150000000\textwidth}
    \centering
    \includegraphics[width=\textwidth]{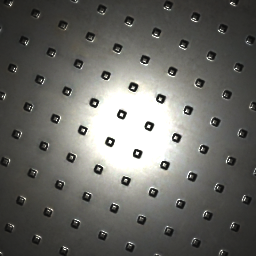}
\end{subfigure}
\begin{subfigure}[b]{0.150000000\textwidth}
    \centering
    \includegraphics[width=\textwidth]{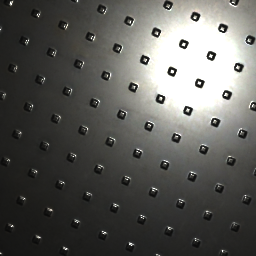}
\end{subfigure}
\begin{subfigure}[b]{0.150000000\textwidth}
    \centering
    \includegraphics[width=\textwidth]{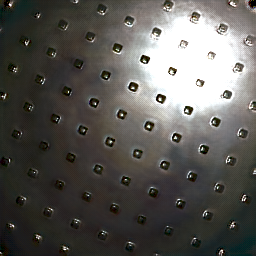}
\end{subfigure}
\begin{subfigure}[b]{0.150000000\textwidth}
    \centering
    \includegraphics[width=\textwidth]{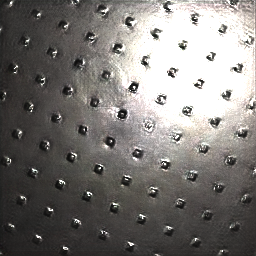}
\end{subfigure}
\begin{subfigure}[b]{0.150000000\textwidth}
    \centering
    \includegraphics[width=\textwidth]{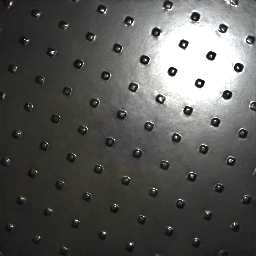}
\end{subfigure}
\begin{subfigure}[b]{0.150000000\textwidth}
    \centering
    \includegraphics[width=\textwidth]{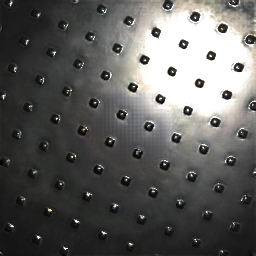}
\end{subfigure}
\centering
\begin{subfigure}[b]{0.150000000\textwidth}
    \centering
    \includegraphics[width=\textwidth]{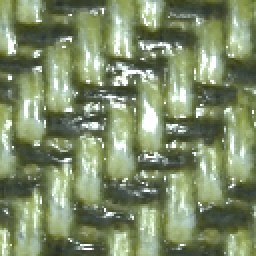}
\end{subfigure}
\begin{subfigure}[b]{0.150000000\textwidth}
    \centering
    \includegraphics[width=\textwidth]{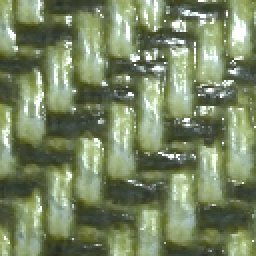}
\end{subfigure}
\begin{subfigure}[b]{0.150000000\textwidth}
    \centering
    \includegraphics[width=\textwidth]{results/ximage_3_row_2.png}
\end{subfigure}
\begin{subfigure}[b]{0.150000000\textwidth}
    \centering
    \includegraphics[width=\textwidth]{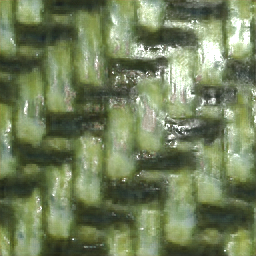}
\end{subfigure}
\begin{subfigure}[b]{0.150000000\textwidth}
    \centering
    \includegraphics[width=\textwidth]{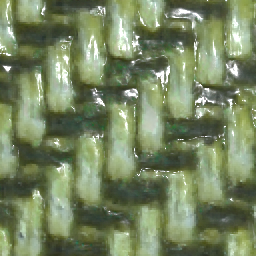}
\end{subfigure}
\begin{subfigure}[b]{0.150000000\textwidth}
    \centering
    \includegraphics[width=\textwidth]{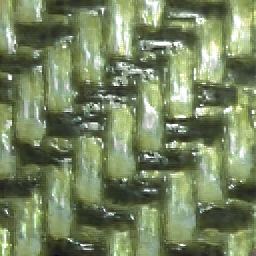}
\end{subfigure}
\centering
\begin{subfigure}[b]{0.150000000\textwidth}
    \centering
    \includegraphics[width=\textwidth]{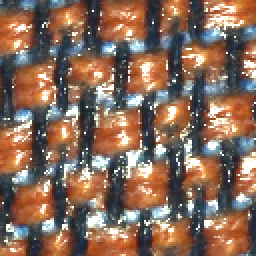}
\end{subfigure}
\begin{subfigure}[b]{0.150000000\textwidth}
    \centering
    \includegraphics[width=\textwidth]{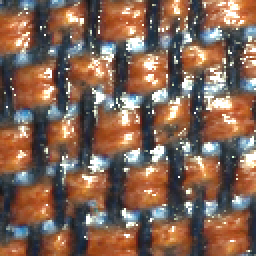}
\end{subfigure}
\begin{subfigure}[b]{0.150000000\textwidth}
    \centering
    \includegraphics[width=\textwidth]{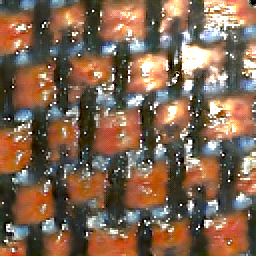}
\end{subfigure}
\begin{subfigure}[b]{0.150000000\textwidth}
    \centering
    \includegraphics[width=\textwidth]{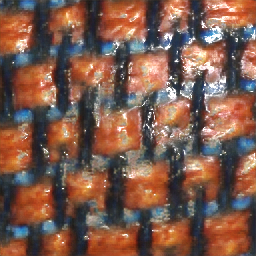}
\end{subfigure}
\begin{subfigure}[b]{0.150000000\textwidth}
    \centering
    \includegraphics[width=\textwidth]{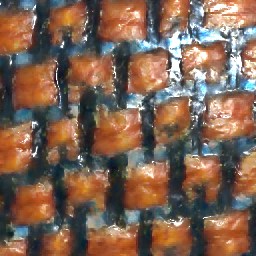}
\end{subfigure}
\begin{subfigure}[b]{0.150000000\textwidth}
    \centering
    \includegraphics[width=\textwidth]{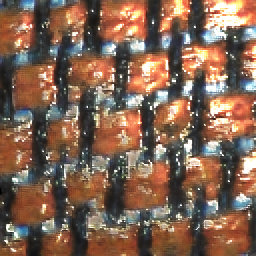}
\end{subfigure}
\centering
\begin{subfigure}[b]{0.150000000\textwidth}
    \centering
    \includegraphics[width=\textwidth]{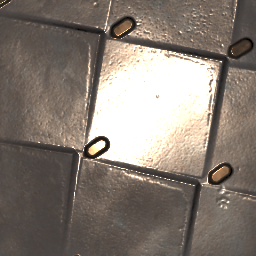}
\end{subfigure}
\begin{subfigure}[b]{0.150000000\textwidth}
    \centering
    \includegraphics[width=\textwidth]{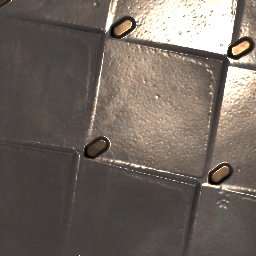}
\end{subfigure}
\begin{subfigure}[b]{0.150000000\textwidth}
    \centering
    \includegraphics[width=\textwidth]{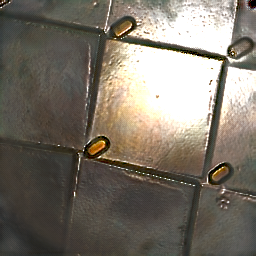}
\end{subfigure}
\begin{subfigure}[b]{0.150000000\textwidth}
    \centering
    \includegraphics[width=\textwidth]{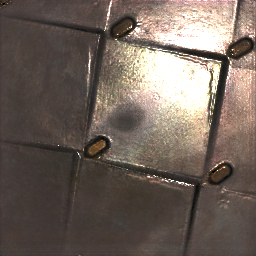}
\end{subfigure}
\begin{subfigure}[b]{0.150000000\textwidth}
    \centering
    \includegraphics[width=\textwidth]{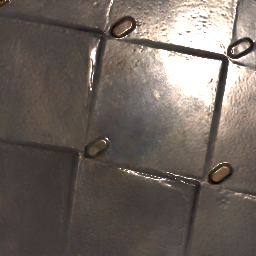}
\end{subfigure}
\begin{subfigure}[b]{0.150000000\textwidth}
    \centering
    \includegraphics[width=\textwidth]{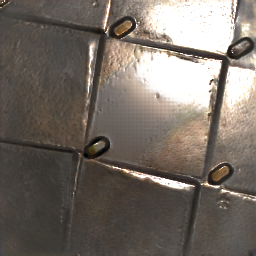}
\end{subfigure}
\centering
\begin{subfigure}[b]{0.150000000\textwidth}
    \centering
    \includegraphics[width=\textwidth]{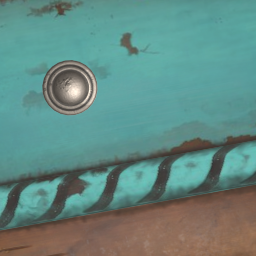}
\end{subfigure}
\begin{subfigure}[b]{0.150000000\textwidth}
    \centering
    \includegraphics[width=\textwidth]{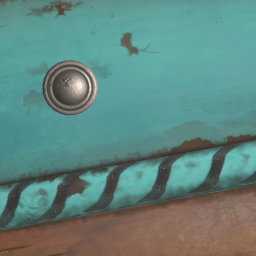}
\end{subfigure}
\begin{subfigure}[b]{0.150000000\textwidth}
    \centering
    \includegraphics[width=\textwidth]{results/ximage_6_row_2.png}
\end{subfigure}
\begin{subfigure}[b]{0.150000000\textwidth}
    \centering
    \includegraphics[width=\textwidth]{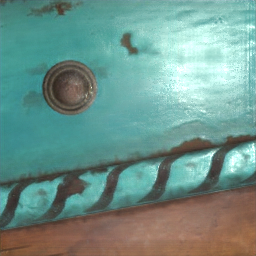}
\end{subfigure}
\begin{subfigure}[b]{0.150000000\textwidth}
    \centering
    \includegraphics[width=\textwidth]{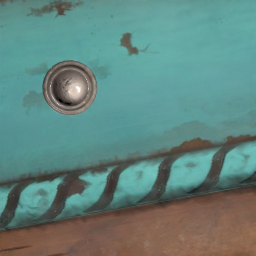}
\end{subfigure}
\begin{subfigure}[b]{0.150000000\textwidth}
    \centering
    \includegraphics[width=\textwidth]{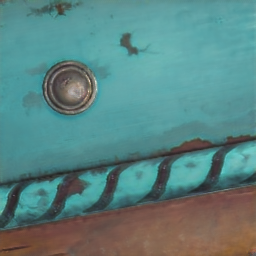}
\end{subfigure}
\centering
\begin{subfigure}[b]{0.150000000\textwidth}
    \centering
    \includegraphics[width=\textwidth]{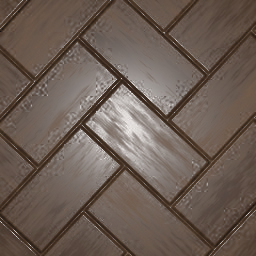}
\end{subfigure}
\begin{subfigure}[b]{0.150000000\textwidth}
    \centering
    \includegraphics[width=\textwidth]{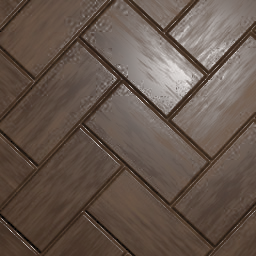}
\end{subfigure}
\begin{subfigure}[b]{0.150000000\textwidth}
    \centering
    \includegraphics[width=\textwidth]{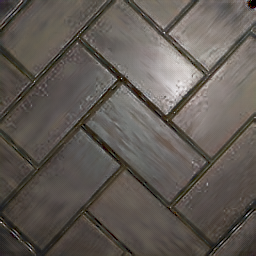}
\end{subfigure}
\begin{subfigure}[b]{0.150000000\textwidth}
    \centering
    \includegraphics[width=\textwidth]{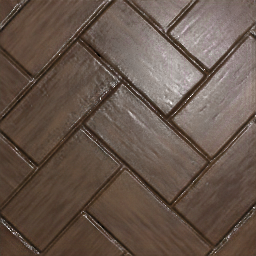}
\end{subfigure}
\begin{subfigure}[b]{0.150000000\textwidth}
    \centering
    \includegraphics[width=\textwidth]{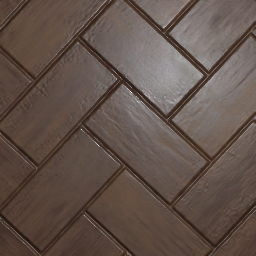}
\end{subfigure}
\begin{subfigure}[b]{0.150000000\textwidth}
    \centering
    \includegraphics[width=\textwidth]{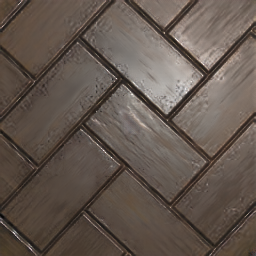}
\end{subfigure}
\caption{Qualitative Comparison of appearance estimation on synthetic photographs. All images are re-rendered with light source translated 1 unit up and 1 unit to the right from the perspective of the camera (centered at [0, 0, 4]).}
\label{fig:synth_demo}
\end{figure*}

\textbf{Comparison To Prior Work.} Figure \ref{fig:synth_demo} compares our neural material estimation technique against the adversarial direct inference method of Zhou and Kalantari \cite{zhou2021adversarial}, the co-located diffusion model of MatFusion \cite{sartor2023matfusion}, and Neural Relighting \cite{bieron2023single}. Qualitatively, our method is comparable to these other recent methods, albeit  does not perform nearly quite as well. This is supported by Table \ref{fig:comp_table}, which tabulates the average perception-based LPIPS render loss \cite{zhang2018unreasonable} over all 50 testing materials for three render cases. \textit{Reflect} renders each exemplar with light and view conditions sampled as specified in our training section.  \textit{Identity} measures how effectively the model 'recreates' the input image given the same light and view as the input photograph. Finally, \textit{Hemisphere} render each exemplar over a set of 128 independently and uniformly sampled point lights and view positions on the hemisphere with radius 4.0 centered at the middle of the material, which matches the FOV of $28^\circ$ used for training of our method and Bieron's Neural Relighting.

While RMSE errors are common for comparing the SVBRDF property maps, our method does not generate analytical SVBRDF maps and so this comparison is not applicable.\\

Due to our method's similarities to Bieron et al.'s \cite{bieron2023single} which performs well here, we posit our slight under-performance is due to poor hyper-parameter selection, especially in regards to the number of positional encoding frequencies and degree of positional encoding compression; this encoding can introduce banded noise from the repetition of frequency signals.  Additionally, the aforementioned instability of $M_{\text{est}}$ results in small 'black-holes' where the renderer is provided high-norm, noisy garbage (see the top-left of our result for the first row of Figure \ref{fig:synth_demo}). Investigating more aggressive instability mitigation strategies appears necessary.

\begin{figure}[htbp]
    \begin{tabularx}{0.5\textwidth}{l|XXX}
 & Reflect & Identity & Hemisphere \\
\hline
Ours & 0.273 & 0.092 & 0.239 \\
Zhou 21 & 0.283 & 0.217 & \textbf{0.200} \\
MatFusion & \underline{0.269} & \underline{0.091} & \underline{0.207} \\
Bieron 23 - Relit & \textbf{0.259} & \textbf{0.075} & 0.219 \\

\end{tabularx}
    \caption{LPIPS metric comparison of our model, Zhou and Kalantari's Adversarial Model \cite{zhou2021adversarial}, Matfusion \cite{sartor2023matfusion}, and Neural Relighting \cite{bieron2023single}. Columns correspond to reflection sampling, identity, and random hemisphere sampling. }
    \label{fig:comp_table}
\end{figure}

\textbf{Real Photographs.} We utilize Zhou and Kalantari's real photograph dataset \cite{zhou2022look} to evaluate how our method generalizes to noisy, real-world photographs. These photographs have no ground truth SVBRDFs to compare re-renders to, and there are only sparse alternative photographs under different light and view directions.

Since these real-world photographs are in a low dynamic range, we train another model identically to the HDR model but using input photographs clamped to the LDR. Further, these real-world photographs are taken on a $45^\circ$ FOV camera, while our model is trained on an FOV of $28^\circ$.

\begin{figure*}[htbp]
\begin{subfigure}[t]{0.150000000\textwidth}
    \centering
    Input
    \vspace{0.800em}
\end{subfigure}
\begin{subfigure}[t]{0.150000000\textwidth}
    \centering
    Reference
    \vspace{0.800em}
\end{subfigure}
\begin{subfigure}[t]{0.150000000\textwidth}
    \centering
    Ours - LDR
    \vspace{0.800em}
\end{subfigure}
\begin{subfigure}[t]{0.150000000\textwidth}
    \centering
    Zhou 21
    \vspace{0.800em}
\end{subfigure}
\begin{subfigure}[t]{0.150000000\textwidth}
    \centering
    MatFusion
    \vspace{0.800em}
\end{subfigure}
\begin{subfigure}[t]{0.150000000\textwidth}
    \centering
    Bieron 23 - Relit
    \vspace{0.800em}
\end{subfigure}
\centering
\begin{subfigure}[b]{0.150000000\textwidth}
    \centering
    \includegraphics[width=\textwidth]{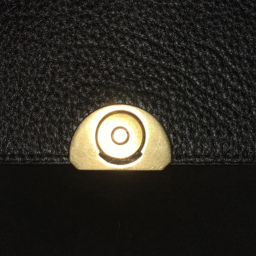}
\end{subfigure}
\begin{subfigure}[b]{0.150000000\textwidth}
    \centering
    \includegraphics[width=\textwidth]{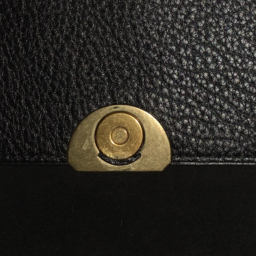}
\end{subfigure}
\begin{subfigure}[b]{0.150000000\textwidth}
    \centering
    \includegraphics[width=\textwidth]{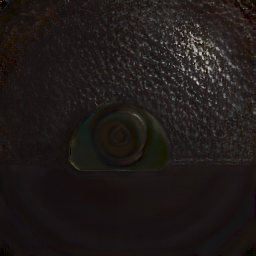}
\end{subfigure}
\begin{subfigure}[b]{0.150000000\textwidth}
    \centering
    \includegraphics[width=\textwidth]{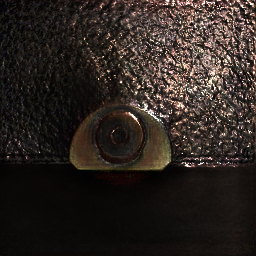}
\end{subfigure}
\begin{subfigure}[b]{0.150000000\textwidth}
    \centering
    \includegraphics[width=\textwidth]{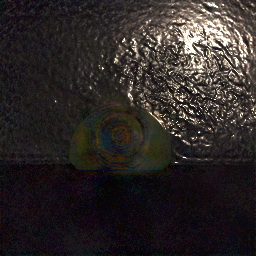}
\end{subfigure}
\begin{subfigure}[b]{0.150000000\textwidth}
    \centering
    \includegraphics[width=\textwidth]{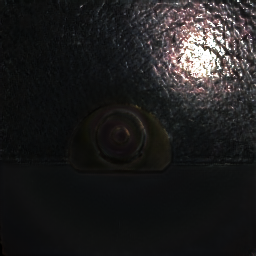}
\end{subfigure}
\centering
\begin{subfigure}[b]{0.150000000\textwidth}
    \centering
    \includegraphics[width=\textwidth]{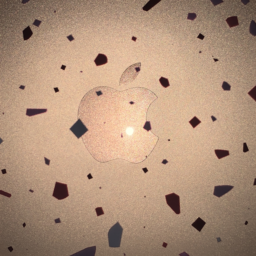}
\end{subfigure}
\begin{subfigure}[b]{0.150000000\textwidth}
    \centering
    \includegraphics[width=\textwidth]{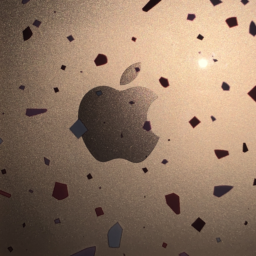}
\end{subfigure}
\begin{subfigure}[b]{0.150000000\textwidth}
    \centering
    \includegraphics[width=\textwidth]{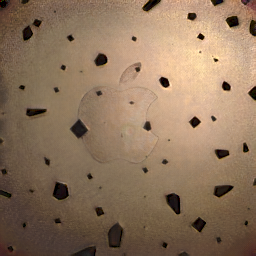}
\end{subfigure}
\begin{subfigure}[b]{0.150000000\textwidth}
    \centering
    \includegraphics[width=\textwidth]{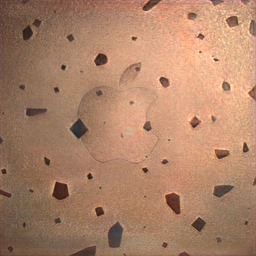}
\end{subfigure}
\begin{subfigure}[b]{0.150000000\textwidth}
    \centering
    \includegraphics[width=\textwidth]{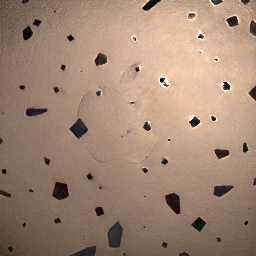}
\end{subfigure}
\begin{subfigure}[b]{0.150000000\textwidth}
    \centering
    \includegraphics[width=\textwidth]{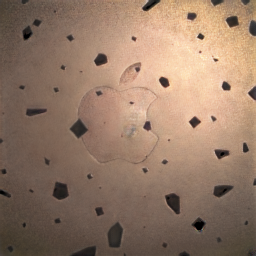}
\end{subfigure}
\centering
\begin{subfigure}[b]{0.150000000\textwidth}
    \centering
    \includegraphics[width=\textwidth]{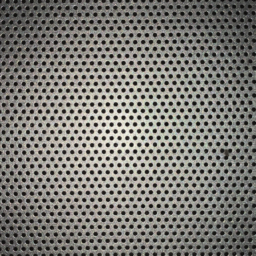}
\end{subfigure}
\begin{subfigure}[b]{0.150000000\textwidth}
    \centering
    \includegraphics[width=\textwidth]{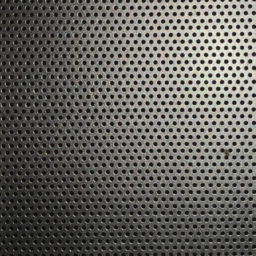}
\end{subfigure}
\begin{subfigure}[b]{0.150000000\textwidth}
    \centering
    \includegraphics[width=\textwidth]{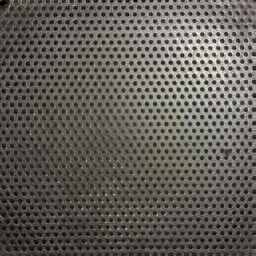}
\end{subfigure}
\begin{subfigure}[b]{0.150000000\textwidth}
    \centering
    \includegraphics[width=\textwidth]{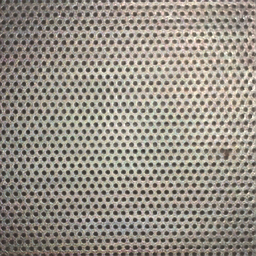}
\end{subfigure}
\begin{subfigure}[b]{0.150000000\textwidth}
    \centering
    \includegraphics[width=\textwidth]{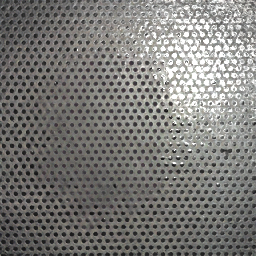}
\end{subfigure}
\begin{subfigure}[b]{0.150000000\textwidth}
    \centering
    \includegraphics[width=\textwidth]{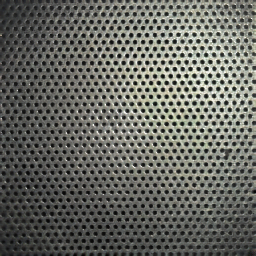}
\end{subfigure}
\centering
\begin{subfigure}[b]{0.150000000\textwidth}
    \centering
    \includegraphics[width=\textwidth]{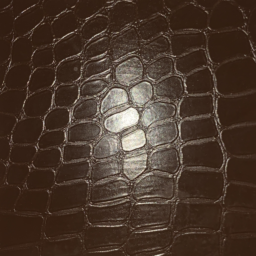}
\end{subfigure}
\begin{subfigure}[b]{0.150000000\textwidth}
    \centering
    \includegraphics[width=\textwidth]{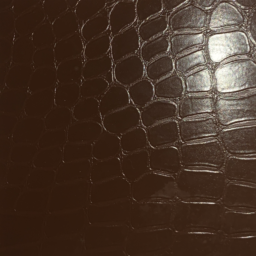}
\end{subfigure}
\begin{subfigure}[b]{0.150000000\textwidth}
    \centering
    \includegraphics[width=\textwidth]{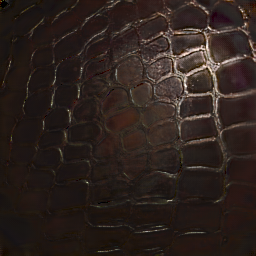}
\end{subfigure}
\begin{subfigure}[b]{0.150000000\textwidth}
    \centering
    \includegraphics[width=\textwidth]{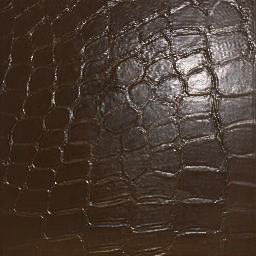}
\end{subfigure}
\begin{subfigure}[b]{0.150000000\textwidth}
    \centering
    \includegraphics[width=\textwidth]{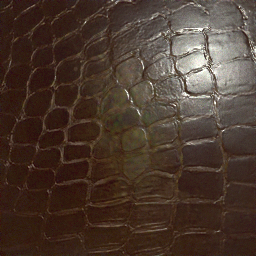}
\end{subfigure}
\begin{subfigure}[b]{0.150000000\textwidth}
    \centering
    \includegraphics[width=\textwidth]{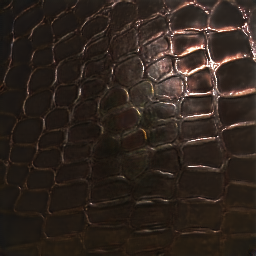}
\end{subfigure}
\centering
\begin{subfigure}[b]{0.150000000\textwidth}
    \centering
    \includegraphics[width=\textwidth]{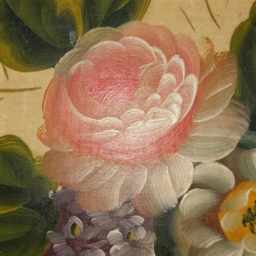}
\end{subfigure}
\begin{subfigure}[b]{0.150000000\textwidth}
    \centering
    \includegraphics[width=\textwidth]{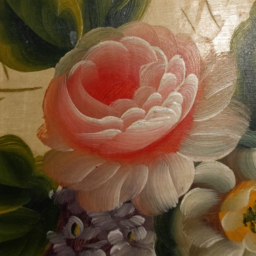}
\end{subfigure}
\begin{subfigure}[b]{0.150000000\textwidth}
    \centering
    \includegraphics[width=\textwidth]{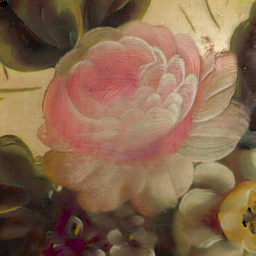}
\end{subfigure}
\begin{subfigure}[b]{0.150000000\textwidth}
    \centering
    \includegraphics[width=\textwidth]{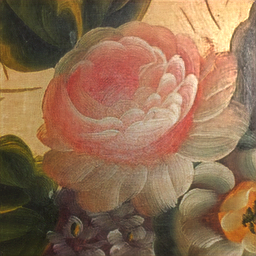}
\end{subfigure}
\begin{subfigure}[b]{0.150000000\textwidth}
    \centering
    \includegraphics[width=\textwidth]{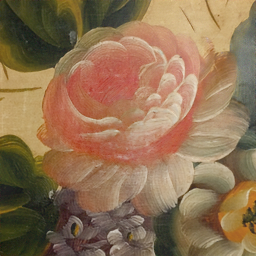}
\end{subfigure}
\begin{subfigure}[b]{0.150000000\textwidth}
    \centering
    \includegraphics[width=\textwidth]{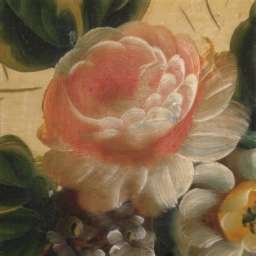}
\end{subfigure}
\centering
\begin{subfigure}[b]{0.150000000\textwidth}
    \centering
    \includegraphics[width=\textwidth]{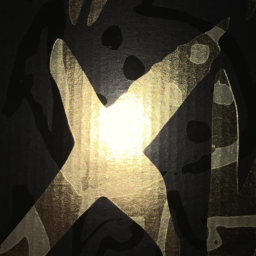}
\end{subfigure}
\begin{subfigure}[b]{0.150000000\textwidth}
    \centering
    \includegraphics[width=\textwidth]{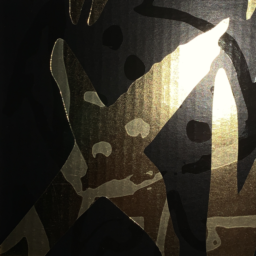}
\end{subfigure}
\begin{subfigure}[b]{0.150000000\textwidth}
    \centering
    \includegraphics[width=\textwidth]{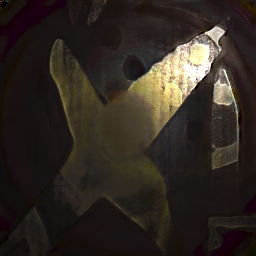}
\end{subfigure}
\begin{subfigure}[b]{0.150000000\textwidth}
    \centering
    \includegraphics[width=\textwidth]{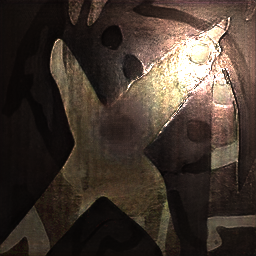}
\end{subfigure}
\begin{subfigure}[b]{0.150000000\textwidth}
    \centering
    \includegraphics[width=\textwidth]{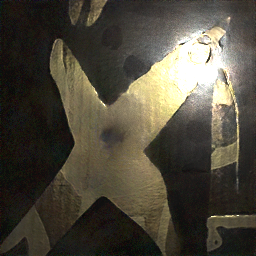}
\end{subfigure}
\begin{subfigure}[b]{0.150000000\textwidth}
    \centering
    \includegraphics[width=\textwidth]{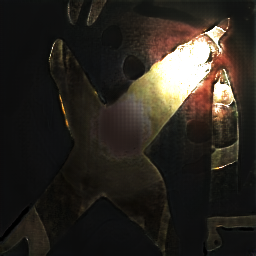}
\end{subfigure}
\centering
\begin{subfigure}[b]{0.150000000\textwidth}
    \centering
    \includegraphics[width=\textwidth]{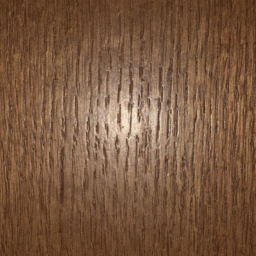}
\end{subfigure}
\begin{subfigure}[b]{0.150000000\textwidth}
    \centering
    \includegraphics[width=\textwidth]{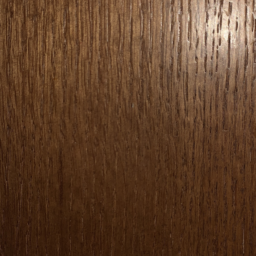}
\end{subfigure}
\begin{subfigure}[b]{0.150000000\textwidth}
    \centering
    \includegraphics[width=\textwidth]{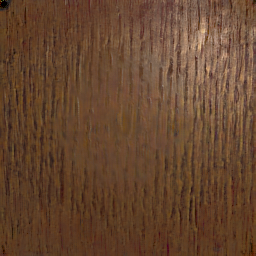}
\end{subfigure}
\begin{subfigure}[b]{0.150000000\textwidth}
    \centering
    \includegraphics[width=\textwidth]{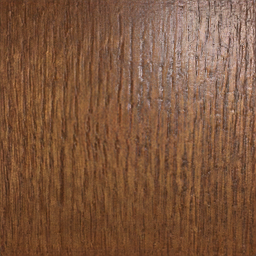}
\end{subfigure}
\begin{subfigure}[b]{0.150000000\textwidth}
    \centering
    \includegraphics[width=\textwidth]{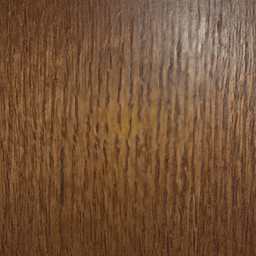}
\end{subfigure}
\begin{subfigure}[b]{0.150000000\textwidth}
    \centering
    \includegraphics[width=\textwidth]{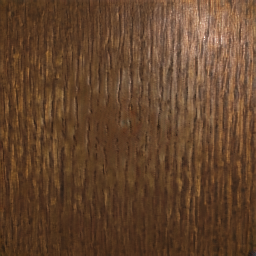}
\end{subfigure}
\centering
\begin{subfigure}[b]{0.150000000\textwidth}
    \centering
    \includegraphics[width=\textwidth]{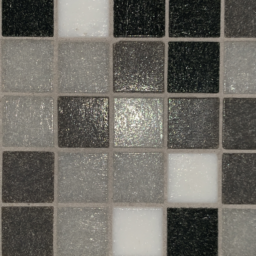}
\end{subfigure}
\begin{subfigure}[b]{0.150000000\textwidth}
    \centering
    \includegraphics[width=\textwidth]{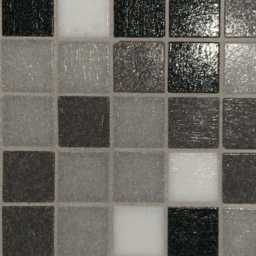}
\end{subfigure}
\begin{subfigure}[b]{0.150000000\textwidth}
    \centering
    \includegraphics[width=\textwidth]{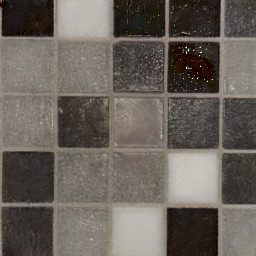}
\end{subfigure}
\begin{subfigure}[b]{0.150000000\textwidth}
    \centering
    \includegraphics[width=\textwidth]{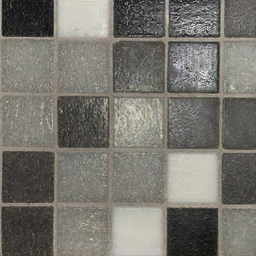}
\end{subfigure}
\begin{subfigure}[b]{0.150000000\textwidth}
    \centering
    \includegraphics[width=\textwidth]{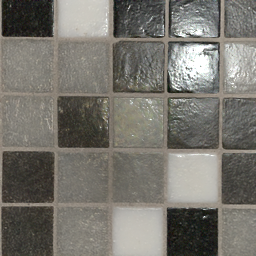}
\end{subfigure}
\begin{subfigure}[b]{0.150000000\textwidth}
    \centering
    \includegraphics[width=\textwidth]{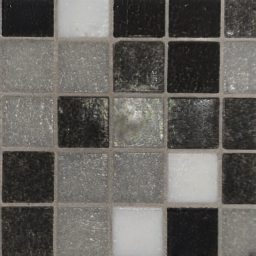}
\end{subfigure}

        \caption{Qualitative Comparison of appearance estimation on real-world LDR photographs with FOV $45^\circ$. Real-world photos sourced from Zhou and Kalantari's real-world photos dataset \cite{zhou2022look}.}
        \label{fig:real_comp}
        \end{figure*}

Figure \ref{fig:real_comp} illustrates how our method performs on these real-world photographs. This demonstrates a limited capacity of our model to generalize to other FOVs and the noise of real photographs. This FOV limitation could be rectified by cropping the input photographs to the expected FOV, or training a model under a different (or variable) FOV.

\begin{figure}[htbp]
    \centering
\begin{subfigure}[t]{0.150000000\textwidth}
    \centering
    No-GI
    \vspace{0.800em}
\end{subfigure}
\begin{subfigure}[t]{0.150000000\textwidth}
    \centering
    GI
    \vspace{0.800em}
\end{subfigure}
\begin{subfigure}[t]{0.150000000\textwidth}
    \centering
    Our Model
    \vspace{0.800em}
\end{subfigure}
\begin{subfigure}[t]{0.470000000\textwidth}
    \includegraphics[width=1\textwidth]{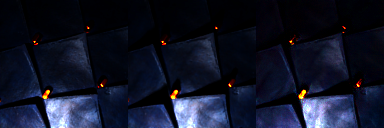} 
\end{subfigure}
\begin{subfigure}[t]{0.470000000\textwidth}
    \includegraphics[width=1\textwidth]{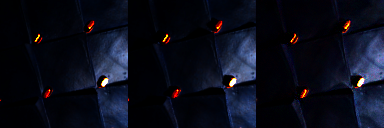} 
\end{subfigure}
\begin{subfigure}[t]{0.470000000\textwidth}
    \includegraphics[width=1\textwidth]{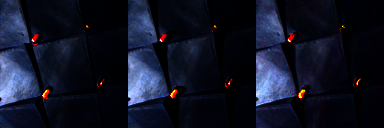} 
\end{subfigure}
    \caption{Qualitative comparison of global illumination transport and shadows, Non-GI render (left), Blender (middle) against our model (right). Note the shadow effect of lower orange nubs.}
    \label{fig:global_demo}
\end{figure}

\subsection{Global Illumination}
Global illumination (GI) transport play a crucial role in creating visually plausible renders. Instead of only accounting for the direct light interaction with points on a surface, GI considers the inter-pixel interactions of light transport. The most obvious instance of GI for near-planar surfaces is shadows, in which a point on a surface block incident light to another point.

Synthetic planar SVBRDF materials are geometrically flat and thus don't natively cast shadows. Perturbations in height (and thus orientation) of a planar material are instead \textbf{simulated} on a per-pixel basis using a normal map to specify the direction perpendicular to the surface at each pixel position. In order for a planar material to cast shadows, its normal map must be integrated to generated a height map, encoding a height component of each pixel; this height map then facilitates simulation of shadows. However, the use of a height map is computationally intensive and requires run-time determination of inter-pixel interactions, which vary widely based on the light position. 

Implicit neural material parameters on the other hand are capable of encoding shadowing information, gleaned from the transformation process from input photograph to neural parameter map. A neural renderer is further capable of simulating shadows using these parameters on a per-pixel basis; no inter-pixel interaction is required by the renderer, only by the first transforming convolutional network.

We demonstrate this capability of our model by \emph{over-fitting} it to a training set rendered using Blender 4, an open-source 3D graphics software, which can produce visualizations of materials with global illumination. Figure \ref{fig:global_demo} demonstrates our model's capability to encode and re-visualize photographs while accounting for global light transport effects.

It is vital to note that due to the highly \emph{over-fitted} training process, $M_{\text{est}}$ cannot be considered as estimating the appearance nor estimating the global light transport of the material, since only a single material is present for training. Instead, we only demonstrate here that shadow information can be encoded into the neural parameters of a pixel and used by a shared neural renderer to simulate global illumination effects.

\subsection{Integration to Path Tracing}
\begin{figure}[htbp]
\begin{subfigure}[t]{0.5\textwidth}
    \centering
\begin{subfigure}[t]{0.32\textwidth}
    \centering
Input
\end{subfigure}
\begin{subfigure}[t]{0.32\textwidth}
    \centering
Neural BRDF
\end{subfigure}
\begin{subfigure}[t]{0.32\textwidth}
    \centering
Ground BRDF
\end{subfigure}
\end{subfigure}
\begin{subfigure}[t]{0.50000000\textwidth}
    \centering
    \includegraphics[width=1.0\textwidth]{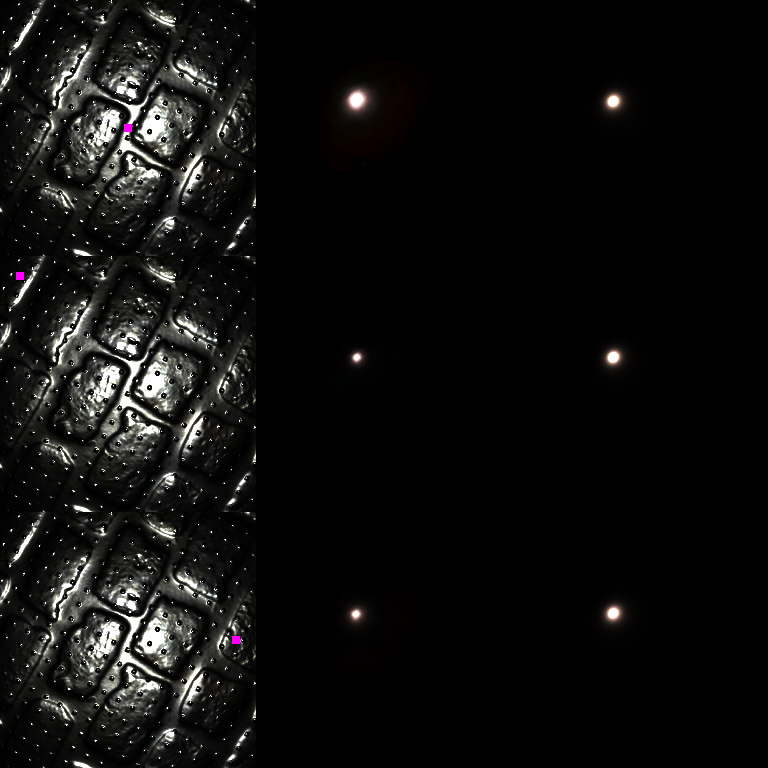} 
    \caption{Glossy, specular bricks BRDF Mitsuba3 example.}
    \label{fig:mitsuba_brick}
\end{subfigure}

\begin{subfigure}[t]{0.50000000\textwidth}
    \centering
    \includegraphics[width=1.0\textwidth]{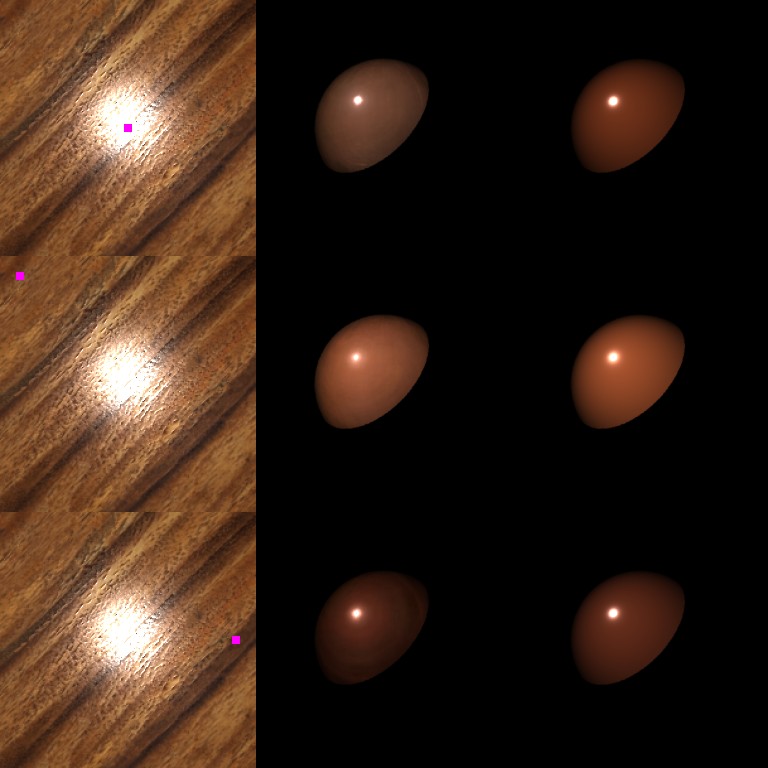} 
    \caption{Glossy and diffuse wood BRDF Mitsuba3 example.}
    \label{fig:mitsuba_wood}
\end{subfigure}
\caption{Individual pixels rendered in Mistuba as independent BRDFs in Mistuba3. Input photograph to model with purple pixel indicating rendered BRDF (Left). Neural BRDF (Middle). Synthetic ground truth BRDF (Right).}
\label{fig:mitsuba}
\end{figure}

$M_{\text{render}}$ combined with a neural parameter vector for a pixel is easily refactored into a BRDF by dividing $M_{\text{render}}$'s output RGB color by the projected area of the pixel of incident light direction. To demonstrate the feasibility of $M_{\text{render}}$ as a practical BRDF, we implement it as a custom BRDF in the Mitsuba3 rendering engine \cite{Mitsuba3}. Figure \ref{fig:mitsuba} shows individual pixels of estimated neural materials rendered as a BRDF applied isotropically to a sphere.

\textbf{Implementation.}\ There are two major implementation concerns: importance sampling and normal correction. 

Importance sampling reduces variance in Monte Carlo rendering by preferentially sampling incident light directions (given a view direction) with higher BRDF reflectance values. The probability density function (PDF) describes the likelihood of sampling each direction given an outgoing view direction and is used to weight importance sampling. While solutions exist to construct a PDF from a neural BRDF using a proxy micro-facet normal distribution PDF \cite{sztrajman2021neural}, we instead opt for a simple diffuse sampling method not adapted to individual BRDFs.

Note that Monte Carlo rendering is not utilized in Figure \ref{fig:mitsuba}, in which only a single infinitesimally small point light is the only light source which can be sampled directly; in scenes with natural light from all directions, a reasonable PDF becomes more relevant for efficient Monte Carlo integration.

Our second concern is normal correction. Every material exemplar is parameterized by a normal map. This normal map encodes perturbations in the surface normals of the material, allowing simulation of surface orientation perturbations without considering actual geometry. Neural parameters encode estimated information on the normal direction of each pixel in order to affects the reflectance behavior of the neural renderer. As such, incident and outgoing directions must be rotated so that this encoded normal is made equivalent to the expected macro-normal of the sphere's surface (we assume the sphere has no normal map). For neural-encoded normal direction, n, and macro-normal m=(0,0,1), this is done via a Gram-Schmidt orthogonal rotation matrix to rotate n to m:

\[
\mathbf{u} = \frac{\mathbf{\mathbf{n} \times \mathbf{r}}}{\|\mathbf{\mathbf{n} \times \mathbf{r}}\|}. \text{\ \ with\ \ \ } \mathbf{r} = [1.0, 0.0, 0.0].\\
\]\[
\mathbf{v} = \mathbf{n} \times \mathbf{u}.
\]\[
\mathbf{R} = \begin{bmatrix}
u_x & v_x & d_x \\
u_y & v_y & d_y \\
u_z & v_z & d_z
\end{bmatrix}.
\] where (x $\times$ y) is the cross product and R is the rotation matrix to rotate input directions by. While inferring n from the encoded neural parameters is possible, we instead choose for simplicity to use the normal encoding in the ground truth SVBRDF parameters, which may be different than the estimated and encoded normal direction of the neural parameters.

\begin{figure}[htbp]
\begin{subfigure}[]{0.15\textwidth}
    \centering
    Ground
\end{subfigure}
\begin{subfigure}[]{0.15\textwidth}
    \centering
    LDR
\end{subfigure}
\begin{subfigure}[]{0.15\textwidth}
    \centering
    HDR
\end{subfigure}

\begin{subfigure}[]{0.15\textwidth}
    \centering
    \includegraphics[width=\textwidth]{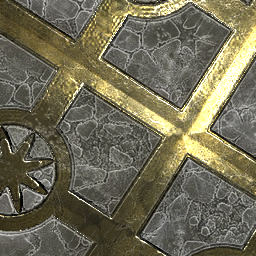}
\end{subfigure}
\begin{subfigure}[]{0.15\textwidth}
    \centering
    \includegraphics[width=\textwidth]{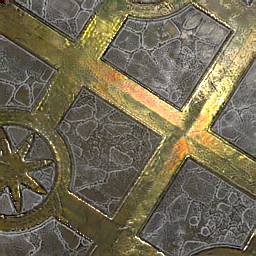}
\end{subfigure}
\begin{subfigure}[]{0.15\textwidth}
    \centering
    \includegraphics[width=\textwidth]{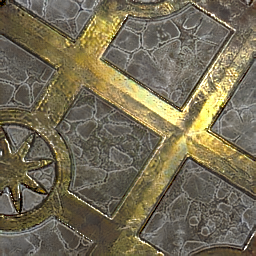}
\end{subfigure}
\begin{subfigure}[]{0.15\textwidth}
    \centering
    \includegraphics[width=\textwidth]{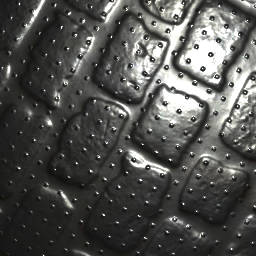}
\end{subfigure}
\begin{subfigure}[]{0.15\textwidth}
    \centering
    \includegraphics[width=\textwidth]{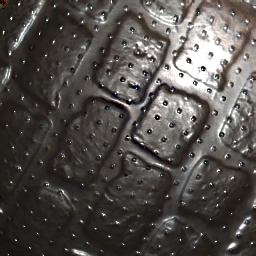}
\end{subfigure}
\begin{subfigure}[]{0.15\textwidth}
    \centering
    \includegraphics[width=\textwidth]{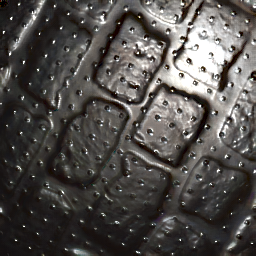}
\end{subfigure}
\begin{subfigure}[]{0.15\textwidth}
    \centering
    \includegraphics[width=\textwidth]{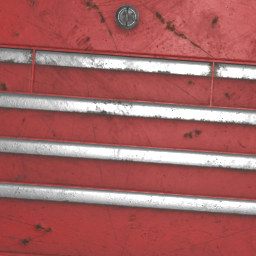}
\end{subfigure}
\begin{subfigure}[]{0.15\textwidth}
    \centering
    \includegraphics[width=\textwidth]{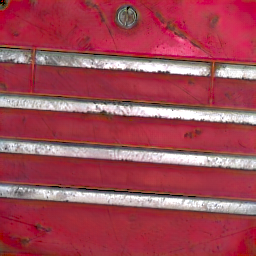}
\end{subfigure}
\begin{subfigure}[]{0.15\textwidth}
    \centering
    \includegraphics[width=\textwidth]{ldr_hdr/re_FullRendersimage_37_row_3.png}
\end{subfigure}

\caption{Qualitative Comparison of LDR and HDR input appearance estimation on real-world photographs with FOV $45^{\circ}$. LDR (middle) and HDR (right).}
\label{fig:ldr_fig}
\end{figure}

\begin{figure}[htbp]
    \begin{tabularx}{0.5\textwidth}{l|XXX}
 & Reflect & Identity & Hemisphere \\
\hline
HDR & 0.273 & 0.092 & \underline{0.239} \\
LDR & \underline{0.270} & \underline{0.085} & 0.242 \\

\end{tabularx}
    \caption{LPIPS metric comparison of our model, Zhou and Kalantari's Adversarial Model \cite{zhou2021adversarial}, Matfusion \cite{sartor2023matfusion}, and Neural Relighting \cite{bieron2023single}. Columns correspond to reflection sampling, identity, and random hemisphere sampling. }
    \label{fig:ldr_table}
\end{figure}

\textbf{Analysis}\ Figure \ref{fig:mitsuba_brick} shows the result of sampling 3 separate pixels from a glossy brick exemplar of the testing set and rendering the estimated neural material parameters and ground truth BRDFs of those pixels onto a sphere in Mitsuba3. Figure \ref{fig:mitsuba_wood} similarly shows the result of sampling 3 separate pixels from a glossy and diffuse wood exemplar of the testing set and rendering their estimated neural material parameters and ground truth BRDFs. Despite predicting the relative location and size of the bright specular highlights correctly, our NBRDFs struggle to fully visualize the smooth variance of diffuse reflectance. Network instability is likely the root cause of the noise and visible banding in the diffuse reflectance of Figure \ref{fig:mitsuba_wood}'s neural renders.

\subsection{Ablation Study}
We perform a simple ablation study to investigate the efficacy of our model to handle the inherent data loss in clamped LDR photograph values. We train two identical architectures on HDR photographs and LDR input photographs separately; all output exemplar renders are still generated as HDR renders for both models. We validate the impact of LDR and HDR input photographs via the same method as the prior work comparison. Figure \ref{fig:ldr_fig} qualitatively shows some key examples of differences in performance between the LDR and HDR models for various materials of the testing dataset of 50 synthetic materials. As with training, the input photographs from the testing set are also clamped to the low dynamic range for the LDR model, while the HDR model's test photographs are not. Table \ref{fig:ldr_table} shows the average perception-based LPIPS render loss \cite{zhang2018unreasonable} over all 50 synthetic testing materials for the same three render cases as Table \ref{fig:comp_table}.

While the LDR model does estimate overall color less accurately in some cases (such as the last row of Figure \ref{fig:ldr_fig}) we see surprisingly minimal variation in the overall LPIPS performance between the models trained on LDR and HDR input photographs. We suspect this is a clear testament to the HA convolution layers' effectiveness within $M_{\text{est}}$.

\section{Conclusion}
We present a novel architecture for single image appearance estimation which foregoes the use of analytical SVBRDF parameters and instead utilizes learned neural material parameters. This project acts as a pivotal initial step towards single image appearance estimation using neural material parameters across the entire re-rendering pipeline. We demonstrated the flexibility of a neural renderer to be implemented in a modern ray-tracing rendering engine, and show the robustness of a neural renderer to light transport effects not captured by single-bounce analytical BRDFs. We would have liked to perform a more in-depth ablation study to investigate the effects of various model architecture features, such as neural parameter count, highlight-aware convolutions, and $M_{\text{render}}$'s architecture. Potential future work may train a two-stage model using the complete MatFusion dataset \cite{sartor2023matfusion} rendered with global illumination to determine if such effects can be estimated; or demonstrate the efficacy of other architectures for neural parameter estimation, such as an Image Transformer architecture \cite{parmar2018image}.

\begin{acks}
This research was supported in part by NSF grant IIS-1909028.
\end{acks}

\bibliographystyle{ACM-Reference-Format}
\bibliography{biblio}

\end{document}